\documentclass[sigconf]{acmart}

\def\BibTeX{{\rm B\kern-.05em{\sc i\kern-.025em b}\kern-.08emT\kern-.1667em\lower.7ex\hbox{E}\kern-.125emX}}
    
\copyrightyear{2018}
\acmYear{2018}
\setcopyright{acmlicensed}
\acmConference[Woodstock '18]{Woodstock '18: ACM Symposium on Neural Gaze Detection}{June 03--05, 2018}{Woodstock, NY}
\acmBooktitle{Woodstock '18: ACM Symposium on Neural Gaze Detection, June 03--05, 2018, Woodstock, NY}
\acmPrice{15.00}
\acmDOI{10.1145/1122445.1122456}
\acmISBN{978-1-4503-9999-9/18/06}

\usepackage{bm}
\begin{document}

%
\title{Gated Group Self-Attention for Answer Selection}

%

\author{Dong Xu$^{1,2}$, Jianhui Ji$^3$, Haikuan Huang$^3$, Hongbo Deng$^3$ and Wu-Jun Li$^{1,2}$}
\affiliation{%
  \institution{$^1$National Key Laboratory for Novel Software Technology, Nanjing University, Nanjing 210023, China}
  \institution{$^2$Department of Computer Science and Technology, Nanjing University, Nanjing 210023, China}
  \institution{$^3$Alibaba Group}
}
\email{xud@lamda.nju.edu.cn,  {jianhui.jjh, haikuan.hhk, dhb167148}@alibaba-inc.com,   liwujun@nju.edu.cn}

%
\renewcommand{\shortauthors}{Trovato and Tobin, et al.}

%
\begin{abstract}
    Answer selection~(answer ranking) is one of the key steps in many kinds of question answering~(QA) applications, where deep models have achieved state-of-the-art performance. Among these deep models, recurrent neural network~(RNN) based models are most popular, typically with better performance than convolutional neural network~(\mbox{CNN}) based models. Nevertheless, it is difficult for RNN based models to capture the information about long-range dependency among words in the sentences of questions and answers. In this paper, we propose a new deep model, called \underline{g}ated \underline{g}roup \underline{s}elf-\underline{a}ttention~(GGSA), for answer selection. \mbox{GGSA} is inspired by global self-attention which is originally proposed for machine translation and has not been explored in answer selection. \mbox{GGSA} tackles the problem of global self-attention that local and global information cannot be well distinguished. Furthermore, an interaction mechanism between questions and answers is also proposed to enhance GGSA by a residual structure. Experimental results on two popular QA datasets show that GGSA can outperform existing answer selection models to achieve state-of-the-art performance. Furthermore, GGSA can also achieve higher accuracy than global self-attention for the answer selection task, with a lower computation cost.
\end{abstract}

\maketitle

\section{Introduction}
\noindent Question answering~(QA) is an important but challenging task in natural language processing~(NLP) area, with wide applications in industry such as intelligent online customer service and intelligent assistant. Answer selection~(answer ranking) is one of the key steps in many kinds of QA applications. For example, in frequently asked questions~(FAQ) applications, models need to select answers from the answer pool that can answer the users' questions. In community-based question answering~(CQA) applications, models need to select the best answer from a set of candidate answers.
  
  Deep neural networks~(DNN) based models, also called deep models, have achieved promising performance for answer selection in recent years~\cite{feng2015applying,tan2016lstm,tay2018cross,tran2018multihop}. Compared with traditional shallow~(non-deep) models~\cite{graber2012besting,riloff2012a,cui2015question}, deep models have several advantages. For example, deep models can automatically extract complex features, while shallow models typically need manually designed features. Deep models can capture semantic features, while shallow models are based on surface lexical features. Furthermore, deep models can achieve better accuracy than shallow models in many QA tasks. Although convolutional neural network~(CNN) based deep models~\cite{feng2015applying} have also been used for answer selection, the most popular deep models for answer selection~\cite{wan2016deep,tan2016lstm,tay2018cross,tran2018multihop} are based on recurrent neural networks~(RNN) and its variants including long-short term memory~(LSTM)~\cite{hochreiter1997long} and gated recurrent units~(GRU)~\cite{cho2014learning}. Both input and computational graph of RNN are sequential, which makes it applicable to model both questions and answers in QA tasks due to their sequential structure. But it is still difficult for existing RNN models to learn long-range dependencies due to the long dependency path that forward and backward signals have to traverse in the network of \mbox{RNN}~\cite{vaswani2017attention}. Hence, existing RNN based models may not well capture the information about long-range dependency among words in the sentences of questions and answers.
  
  Recently, a new deep model, called Transformer~\cite{vaswani2017attention}, is proposed for sequence modeling. \mbox{Transformer} discards CNN and RNN structure entirely, using only stacked self-attention\footnote{The self-attention methods used in Transformer~\cite{vaswani2017attention} and RNN based models~\cite{shen2017inter,cheng2016long,lin2017a,parikh2016a} are different. In the context of this paper, self-attention refers to the self-attention mechanism with multi-head attention and residual structure, which is proposed in \cite{vaswani2017attention}. While intra-attention refers to the self-attention method used in RNN based models~\cite{shen2017inter,cheng2016long,lin2017a,parikh2016a}. } for modeling. In self-attention, the maximum dependency path length between two words within a sequence is one, which makes self-attention learn long-range dependency in a sequence better than RNN. Self-attention is originally proposed for machine translation, showing better performance than CNN and RNN~\cite{vaswani2017attention}. Furthermore, self-attention has also proved its effectiveness in other tasks including video classification~\cite{wang2018non}, reading comprehension~\cite{wei2018fast}, semantic role labeling~\cite{tan2018deep}, and recommendation~\cite{zhou2018atrank}. However, for the answer selection task, there have not existed self-attention based models. Furthermore, the self-attention adopted in existing models is actually \emph{global self-attention} which means that the meaning~(semantic) of a word is determined by the meaning of all words in the sequence in a global way, without distinction between its surrounding words~(local information) and all words in the sequence~(global information).
  
  In this paper, we propose a new self-attention based deep model, called \underline{g}ated \underline{g}roup \underline{s}elf-\underline{a}ttention~(GGSA), for answer selection. The contributions of GGSA are briefly outlined as follows:
  \begin{itemize}
    \item \mbox{GGSA} tackles the problem of global self-attention that local and global information cannot be well distinguished. More specifically, GGSA distinguishes the local information and global information through \emph{group self-attention}. And then these two different kinds of information are combined through a gate mechanism.
    \item To the best of our knowledge, GGSA is the first self-attention based model for answer selection.
    \item In GGSA, a novel interaction mechanism between questions and answers is also proposed through a residual structure. The interaction mechanism calculates question-aware residuals which are then added to the representation of words in answers, which can make the words in answers to take question information into account.
    \item Experimental results on two popular QA datasets show that GGSA can outperform existing answer selection models to achieve state-of-the-art performance. Furthermore, GGSA can also achieve higher accuracy than global self-attention  for the answer selection task, with a lower computation cost.
  \end{itemize}
  
  The following content of this paper is organized as follows. Section 2 briefly introduces the self-attention structure proposed by \cite{vaswani2017attention}. The details of GGSA which we proposed for answer selection are presented in Section 3. Section 4 describes the experimental results. The related works are discussed in Section 5. Finally, we conclude and discuss future work in Section 6.

  \section{Self-Attention}
  Self-attention is an attention mechanism that the attention weights are generated by a single sequence itself to compute the sequence representation. It is also known as intra-attention which is generally adopted in RNN structures~\cite{shen2017inter,cheng2016long,lin2017a,parikh2016a}. Recently, a new self-attention~\cite{vaswani2017attention} structure has been proposed and has achieved remarkable results in a wide range of tasks~\cite{wang2018non,wei2018fast,zhou2018atrank}. For distinguishing these two self-attention methods in this paper, self-attention refers to the self-attention method with multi-head attention and residual structure, which is proposed in \cite{vaswani2017attention}. While intra-attention refers to the self-attention method which is adopted in RNN structures~\cite{shen2017inter,cheng2016long,lin2017a,parikh2016a}.
  
  GGSA is inspired by self-attention~\cite{vaswani2017attention}, which is also called \emph{global self-attention} in this paper. Here, we give a brief introduction about global self-attention, including self-attention block and positional encoding.
  
  \subsection{Self-Attention Block}
\begin{figure}[htb]
    \centering
    \includegraphics[width=0.25\textwidth]{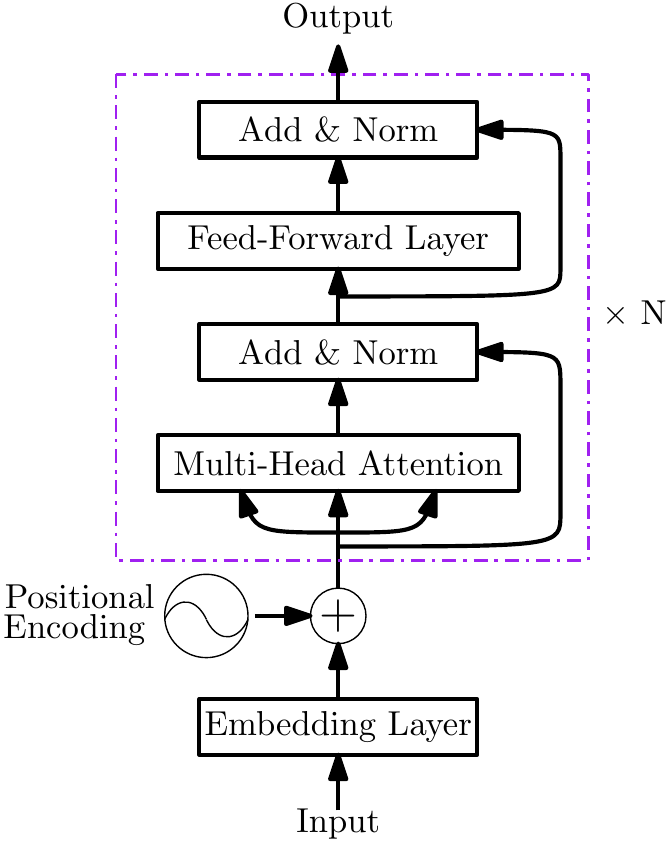}
    \caption{\label{self-attention-block}The architecture of a self-attention block.}
  \end{figure}

  The major component of self-attention based deep models, such as Transformer~\cite{vaswani2017attention}, is a stack of several self-attention blocks. A self-attention block~\cite{vaswani2017attention} consists of a multi-head attention module, a position-wise feed-forward layer and layer normalization~\cite{ba2016layer}. These elements are constructed into a residual network structure which is shown in Figure~\ref{self-attention-block}.
  
  The input to a self-attention block is a sequence of words. In this paper, we use word embedding to represent the words. Suppose the dimension of word embedding is $D$, sequence length is $L$, the input~(sequence) is represented by a matrix $\bm{X} \in \mathbb{R}^{D \times L}$. The output $\bm{H} \in \mathbb{R}^{D \times L}$, which denotes the learned representation for the words in the input sequence by the self-attention block, can be defined as follows:
  \begin{gather}
    \bm{C} = MultiHeadAttention(X), \nonumber \\
    \bm{Y} = LayerNorm(\bm{X} + \bm{C}),  \nonumber  \\
    \bm{R} = FeedForwardLayer(\bm{Y}),  \nonumber  \\
    \bm{H} = LayerNorm(\bm{Y} + \bm{R}),  \nonumber
  \end{gather}
  where $\bm{C}$ and $\bm{R}$ are the residuals calculated by multi-head attention and feed-forward layer respectively. $\bm{Y}$ is the normalized intermediate output of the sequence. $FeedForwardLayer(*)$ denotes the operation of a two-layer feed-forward network~(FFN) with ReLU activation and $LayerNorm(*)$ denotes the layer normalization proposed by \cite{ba2016layer}.
  
  The following content describes the $MultiHeadAttention$ in detail, which is the key module of self-attention. In order to use attention, the input $\bm{X}$ is first transformed into query, key, value through linear operations. The output of attention is the weighted sum of values, where the weights are computed by a scaled dot-product of query and key. Finally, a linear operation is applied to the output. The formulation is as follows:
  \begin{gather}
    [\bm{Q}, \bm{K}, \bm{V}] = [\bm{W}^q, \bm{W}^k, \bm{W}^v] \cdot \bm{X}, \nonumber \\
    Attention(\bm{Q}, \bm{K}, \bm{V}) = \bm{V} \cdot Softmax(\frac{\bm{Q}^\top \cdot \bm{K}}{\sqrt{D}}), \nonumber \\
    \bm{C} = \bm{W}^o \cdot Attention(\bm{Q}, \bm{K}, \bm{V}). \nonumber
  \end{gather}
  where $\bm{Q}$, $\bm{K}$ and $\bm{V}$ denote query, key and value, respectively. $\bm{W}^q,$ $\bm{W}^k, \bm{W}^v, \bm{W}^o \in \mathbb{R}^{D \times D}$ are the parameters.
  
  The above formulas are for a special case when the number of heads equals to $1$. The model can have $n$ heads. In this case, $\bm{Q}, \bm{K}, \bm{V}$ are divided into $n$ parts~(subspaces), denoted as $\{\bm{Q}_i, \bm{K}_i, \bm{V}_i|i=1,2,\cdots,n\}$ where $\bm{Q}_i, \bm{K}_i, \bm{V}_i \in \mathbb{R}^{d \times L}$ and $d = D / n$. This is equivalent to $\bm{W}^q, \bm{W}^k, \bm{W}^v$ being divided into $n$ parts~(subspaces), denoted as $\{\bm{W}^q_i, \bm{W}^k_i, \bm{W}^v_i | i=1,2,\cdots,n\}$ where $\bm{W}^q_i, \bm{W}^k_i, \bm{W}^v_i \in \mathbb{R}^{d \times D}$ and $d = D / n$. $\bm{W}^q_i, \bm{W}^k_i, \bm{W}^v_i$ are used to transform $\bm{X}$ into the $i$-th subspaces $\bm{Q}_i, \bm{K}_i, \bm{V}_i$. Then $n$ times of attention are performed and the outputs are concatenated. Finally, an affine operation $\bm{W}^o \in \mathbb{R}^{D \times D}$ is used to transform $n$ subspaces back to the original space and get the representation, which is formulated as follows:
  \begin{gather}
    \bm{S}_i = Attention(\bm{Q}_i, \bm{K}_i, \bm{V}_i), \nonumber \\
    \bm{C} = \bm{W}^o \cdot Concat(\bm{S}_1, ..., \bm{S}_n). \nonumber
  \end{gather}
  The motivation of multi-head attention is that it allows the model to jointly attend to different context from different subspaces~\cite{vaswani2017attention}.
  
\begin{figure}[htb]
    \centering
    \includegraphics[width=0.25\textwidth]{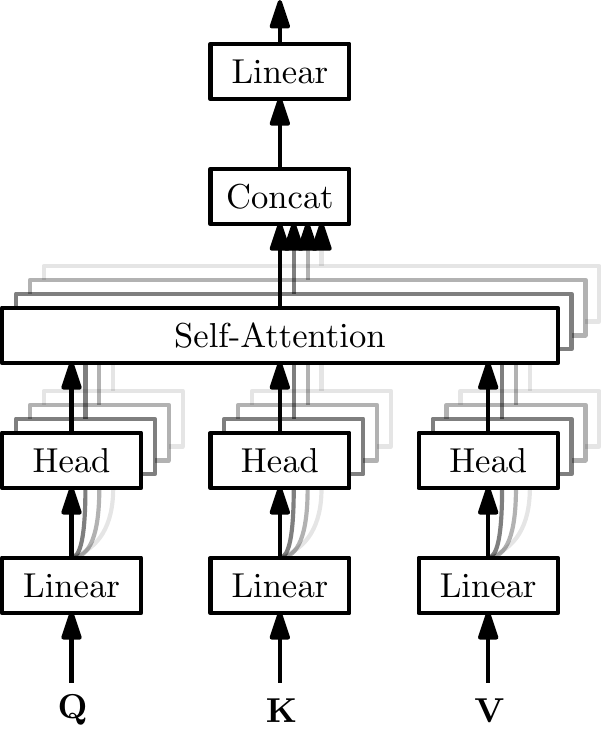}
    \caption{\label{multi-head-attention}Multi-head attention when the number of heads equals to four, which means that the representations~($\bm{Q}, \bm{K} ,\bm{V}$) of each word is split into four parts~(subspaces).}
  \end{figure}

  Figure~\ref{multi-head-attention} illustrates an example of multi-head attention when the number of heads equals to four.
  \subsection{Positional Encoding}
  \noindent Since the attention mechanism ignores the order of sequence, \cite{vaswani2017attention} adds positional encoding to the input embeddings of the words.
  \begin{gather}
    PE(i,j) =
    \begin{cases}
      sin(i/10000^{2j/D})& \text{$j$ is even} \nonumber \\
      cos(i/10000^{2j/D})& \text{$j$ is odd}
    \end{cases}
  \end{gather}
  where $i$ is the position of word in the sequence and $j$ represents the $j$-th dimension of embedding. This positional encoding is more simple and effective than the trainable positional embedding in~\cite{gehring2017convolutional}, with nearly identical accuracy. Hence, we also use this positional encoding in our model.

  \section{Gated Group Self-Attention}
  In this section, we present the details of our proposed model called \underline{g}ated \underline{g}roup \underline{s}elf-\underline{a}ttention~(GGSA). GGSA has two versions: GGSA and GGSA with \underline{i}nteraction~(iGGSA). Both GGSA and iGGSA use the same gated group self-attention structure except that iGGSA uses an additional interaction mechanism to incorporate question information into the encoding process of answer. We will first describe these two versions of GGSA respectively. And then, we will introduce several upper structures~(networks) which can adopt GGSA or iGGSA as a key building block.
  
  \subsection{GGSA}
  We will first introduce group self-attention which can distinguish local information and global information in a sequence, and then introduce a gate mechanism which can combine these two different kinds of information.
  
  \subsubsection{Group Self-Attention}
\begin{figure}[t]
    \centering
    \includegraphics[width=0.35\textwidth]{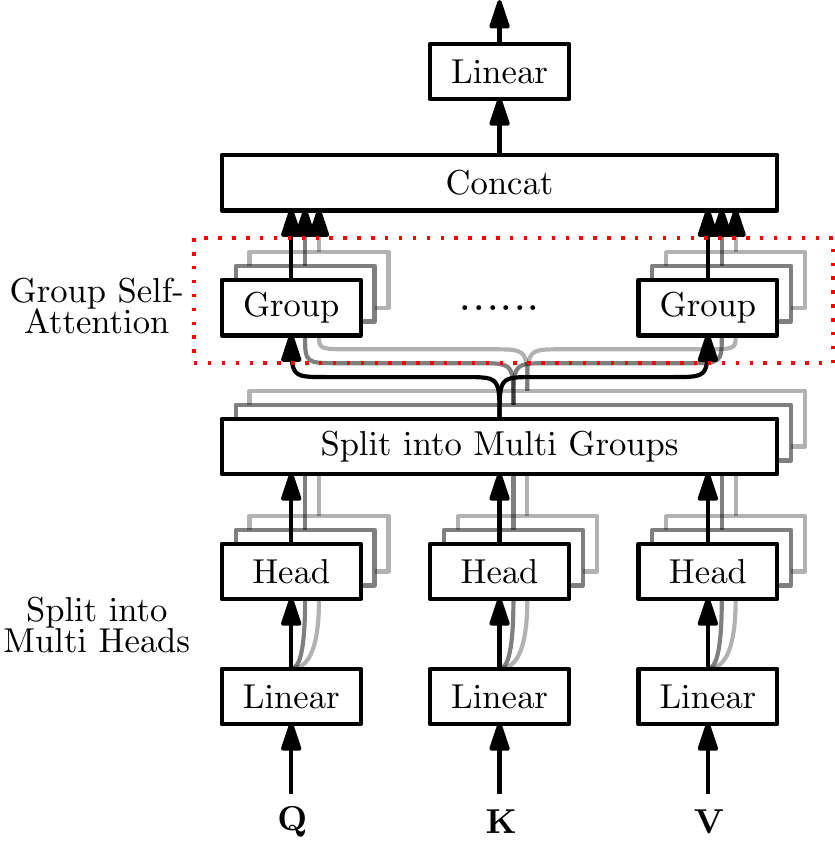}
    \caption{\label{group-multi-head-attention} The structure of group self-attention. After $\bm{Q}, \bm{K}, \bm{V}$ are split into several heads, each head is split into multiple groups.}
  \end{figure}

  The idea of group self-attention is motivated by a fact that, for a word, the local context is more important than the distant context. In other words, the meaning of a word is determined by two parts of information, the meaning of its surrounding words~(local information) and the meaning of all words~(global information). But the difference between these two parts of information is not explicitly reflected in global self-attention. In global self-attention, each word performs attention with all words in the sequence with the same distance one where the difference between local and global information is only implicitly distinguished by positional encoding. Furthermore, every word has to perform attention with all words, which is not always necessary and will lead to huge overhead. Based on the above reasons, we propose the architecture of group self-attention. In group self-attention, the whole sequence is divided into several groups with the same group size, and each word only performs attention with words in the same group. Figure~\ref{group-multi-head-attention} illustrates the structure of group self-attention.
  
  \begin{figure}[tb]
    \centering
    \includegraphics[width=0.35\textwidth]{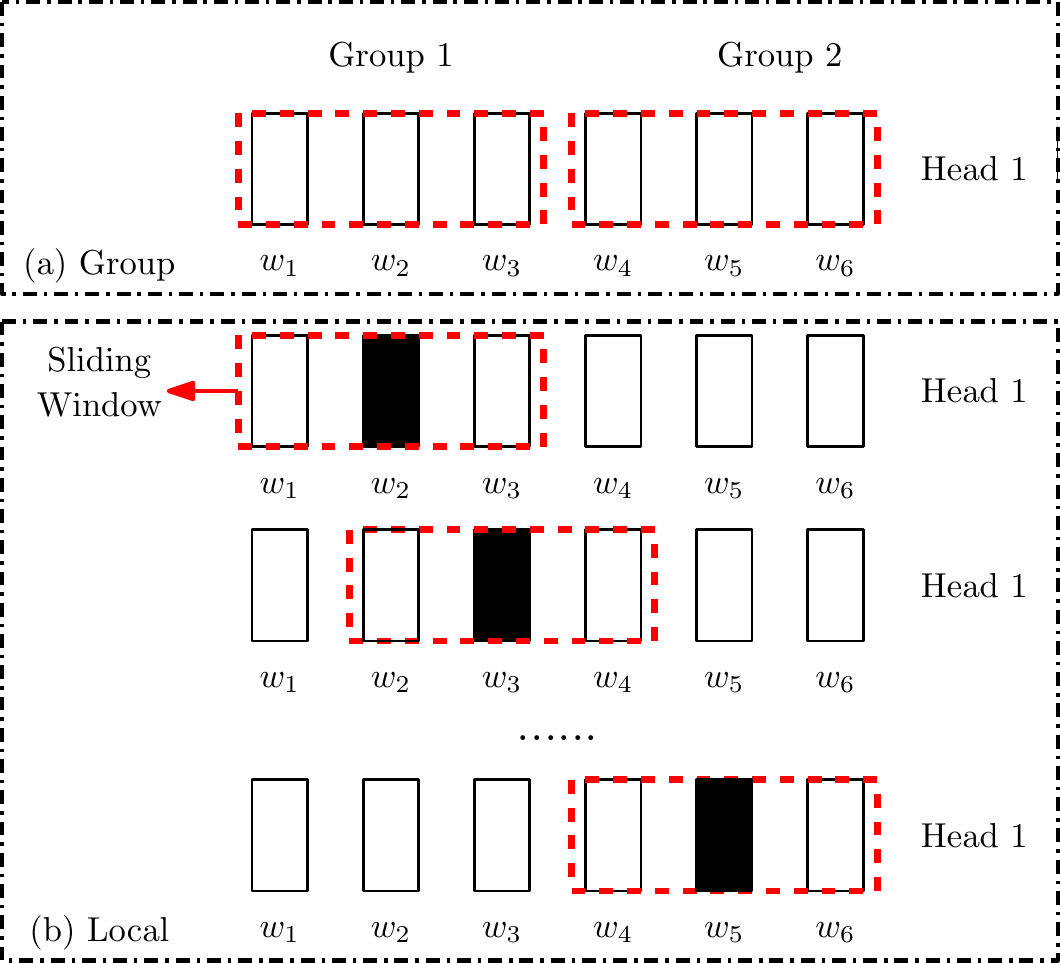}
    \caption{\label{local-vs-group} (a). Group self-attention when the group size equals to three. (b). Local self-attention when the sliding window size equals to three. For a given word $w_i$, it will be split into the same group or sliding window in different heads. For $w_3$ in group self-attention, its receptive fields are $w_1, w_2, w_3$ in each head. For $w_3$ in local self-attention, its receptive fields are $w_2, w_3, w_4$ in each head. So we just show the situation where the number of heads equals to one.}
  \end{figure}
  Group self-attention is different from local self-attention which is illustrated in Figure~\ref{local-vs-group}~(b). In group self-attention, the sequence is divided into different groups that do not overlap with each other. Then, self-attention is performed within groups and the words in the same group has the same receptive field. But in local self-attention, a sliding window moves from left to right with a stepsize being one word and consecutive windows will overlap with each other. In each sliding window, the self-attention is only performed for the central word, which means that the sliding window is actually the receptive field of the central word. So the receptive fields are different for each word in the sequence. Both group self-attention and local self-attention can capture local information and can be applied to GGSA. The reason why we choose group self-attention in our experiments is mainly due to efficiency consideration. Local self-attention needs to use an operation called ``img2col'' which is designed for convolution. This operation needs to expand the original matrix, which increases the overhead of memory and computation.

  Similar to that in global self-attention, we also adopt multiple heads for attention in our group self-attention which is denoted by $GroupMultiHeadAttention$. $\bm{Q}, \bm{K}, \bm{V}$ are first divided into $n \times m$ parts (subspaces) $\{\bm{Q}_{i,j}, \bm{K}_{i,j}, \bm{V}_{i,j}|i=1,2,\cdots,n; j =1,2,\cdots,m\}$, where $n$ and $m$ are the number of heads and groups, respectively. $\bm{Q}_{i,j}\in \mathbb{R}^{d \times l}, \bm{K}_{i,j}\in \mathbb{R}^{d \times l}, \bm{V}_{i,j} \in \mathbb{R}^{d \times l}$, where $d = D / n$, $l = L / m$. Then $n \times m$ times of attention are performed and the $n \times m$ outputs are concatenated:
  \begin{gather}
    \bm{T}_{i,j}= Attention(\bm{Q}_{i,j}, \bm{K}_{i,j}, \bm{V}_{i,j}), \nonumber  \\
    \bm{C} = \bm{W}^o \cdot Concat(\bm{T}_{1,1}, ..., \bm{T}_{n,m}). \nonumber
  \end{gather}
  
  \subsubsection{Offset Strategy}
\begin{figure}[htb]
    \centering
    \includegraphics[width=0.3\textwidth]{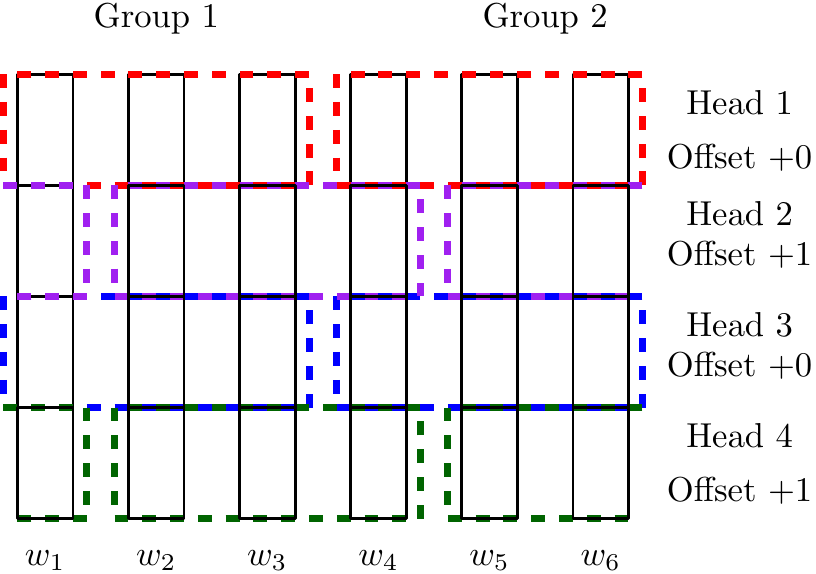}
    \caption{\label{offset-strategy}Offset strategy in group self-attention, with group size being 3 and number of heads being 4.}
  \end{figure}

  Note that the receptive fields of some words are limited after grouping, especially for those words at the boundary of groups. e.g. $w_3$ in Figure \ref{local-vs-group} (a) can not attend to the surrounding word $w_4$, because they are divided into different groups. We design an offset strategy to solve this problem, which is illustrated in Figure~\ref{offset-strategy}. More specifically, we divide the input into multiple heads and different heads might have different offsets. Because of offset, a word may be divided into different groups in different heads, which leads to a larger receptive field for that word. e.g. $w_3$ in Figure \ref{offset-strategy} can not attend to the surrounding word $w_4$ in head $1$, but can attend to $w_4$ in head $2$.
  
  In another view, the offset strategy increases the diversity of different heads~(subspaces). More specically, for a given word, the receptive fields are different in different heads~(subspaces). This makes GGSA attend to different locations in different heads~(subspace) better than other self-attention structures.
  
  With this offset strategy, group self-attention can achieve the same approximate function of local self-attention with a much lower computation cost. So, we adopt group self-attention to construct our GGSA model.
  
  \subsubsection{Gate Mechanism} The group self-attention proposed above can capture local information about the words. In real applications, the information from all words~(global information) is also important for modeling the semantic~(meaning) of a word. In GGSA, we propose a global information gate to combine local information and global information into the same framework. The details of global information gate is illustrated in Figure~\ref{global-info-gate}. To put all modules mentioned before together, the formulations of GGSA can be list as follows:
  \begin{figure}[t]
    \centering
    \includegraphics[width=0.35\textwidth]{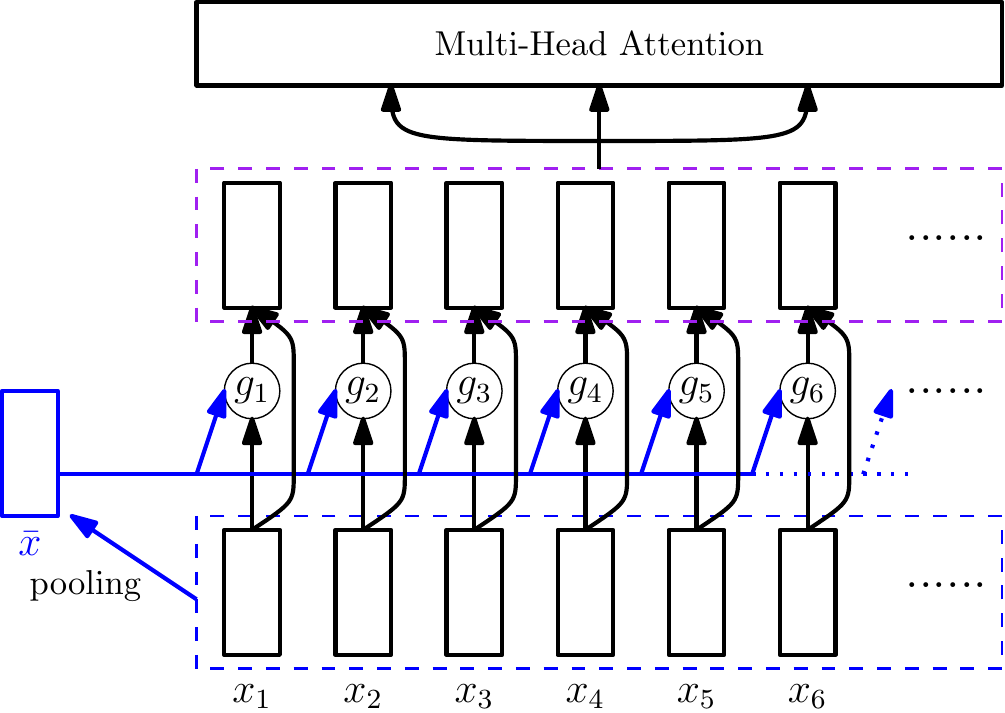}
    \caption{\label{global-info-gate}The structure of global information gate.}
  \end{figure}
  \begin{gather}
    \bar{\bm{x}} = \frac{\sum_i \bm{x}_i}{L}, \label{pooling-x}\\
    \bm{g}_i = \sigma (\bm{W} \cdot (\bm{x}_i \odot \bar{\bm{x}}) + \bm{b}), \nonumber \\
    \bm{C} = GroupMultiHeadAttention(\bm{X} \odot \bm{G}), \nonumber \\
    \bm{Y}= LayerNorm(\bm{X} + \bm{C}), \label{middle} \\
    \bm{R} = FeedForwardLayer(\bm{Y}), \label{GGSAN-5}\\
    \bm{H} = \bm{Y} + \bm{R}, \label{GGSAN-6}
  \end{gather}
  where $\bm{x}_i$ is the representation of word $i$, which is the $i$-th column of $\bm{X}  \in \mathbb{R}^{D \times L}$, $L$ is the length of the sequence. $\bar{\bm{x}}$ denotes the result of a mean~(average) pooling operation on the words of $\bm{X}$, which can be seen as the context vector with global information. $\bm{W} \in \mathbb{R}^{D \times D}$ is the learnable weight and $\bm{b}$ is the bias. $\bm{G} \in \mathbb{R}^{D \times L}$ is the global information gate which will be applied to the input $\bm{X}$, with the \mbox{$i$-th} column of $\bm{G}$ being $\bm{g}_i$. The gated input $\bm{X} \odot \bm{G}$ is passed through group self-attention to compute the residual $\bm{C}$. The left operations in (\ref{middle}), (\ref{GGSAN-5}) and (\ref{GGSAN-6}) are the same as those in global self-attention, except that the last layer normalization is removed in group self-attention because we find it is not necessary in our experiments.
  
  Here, we will show some intensive analyses about the global information gate and the way that it can combine local information and global information. On one hand, when $\bm{g}_i$ is open, it means that the current word $\bm{x}_i$ is more correlated with global information $\bar{\bm{x}}$. When $\bm{g}_i$ is closed, it means that the current word $\bm{x}_i$ is unrelated to global information $\bar{\bm{x}}$. On the other hand, the context vector $\bar{\bm{x}}$ is used to calculate $\bm{g}_i$, thus $\bm{g}_i$ can capture the information of $\bar{\bm{x}}$. If we treat the gate $\bm{g}_i$ as a vector representation, the gate itself contains global information. As a results, $\bm{g}_i$ can pass the global information to the representation of current word through multiplication $\bm{x}_i \odot \bm{g}_i $.
  
  Figure~\ref{GGSAN-i}~(a) illustrates the model architecture of GGSA for answer selection, where figure~\ref{GGSAN-i}~(c) is a GGSA encoder. GGSA uses two GGSA encoders to encode the question and the answer respectively. Then, the outputs of GGSA encoders are passed into composition layer and similarity layer to compute the matching score of the (question, answer) pair.
  \subsection{iGGSA}
  
  The same word in answers may have different meanings corresponding to different questions. For example, when we ask two questions about cell phone and food respectively, the meaning of ``apple'' in the answers for one question is different from that for the other question. Here, we design a question-answer interaction mechanism to model the interaction between a question and an answer. The interaction mechanism calculates a question-aware residual $\widetilde{\bm{R}}$ which is then added to the intermediate output $\bm{Y}^a$ of GGSA in the answer part. With this question-answer interaction mechanism, the representations of words in answers will capture the context information of questions. The effect of this question-answer interaction mechanism is different from the effect of traditional attention mechanism~\cite{tan2016lstm} used in RNN based models. The attention mechanism~\cite{tan2016lstm} plays a role for selecting the related information from the whole answer sequence, while question-answer interaction mechanism calculates a residual which leads the representations of the words in answer sequence to the right meaning in the context of question. So, iGGSA can also adopt attention for composition. In this paper, we use iGGSA to denote the GGSA version with interaction between questions and answers, and use GGSA to denote the version without interaction between questions and answers.
  
  \begin{figure*}[htb]
    \centering
    \includegraphics[width=0.9\textwidth]{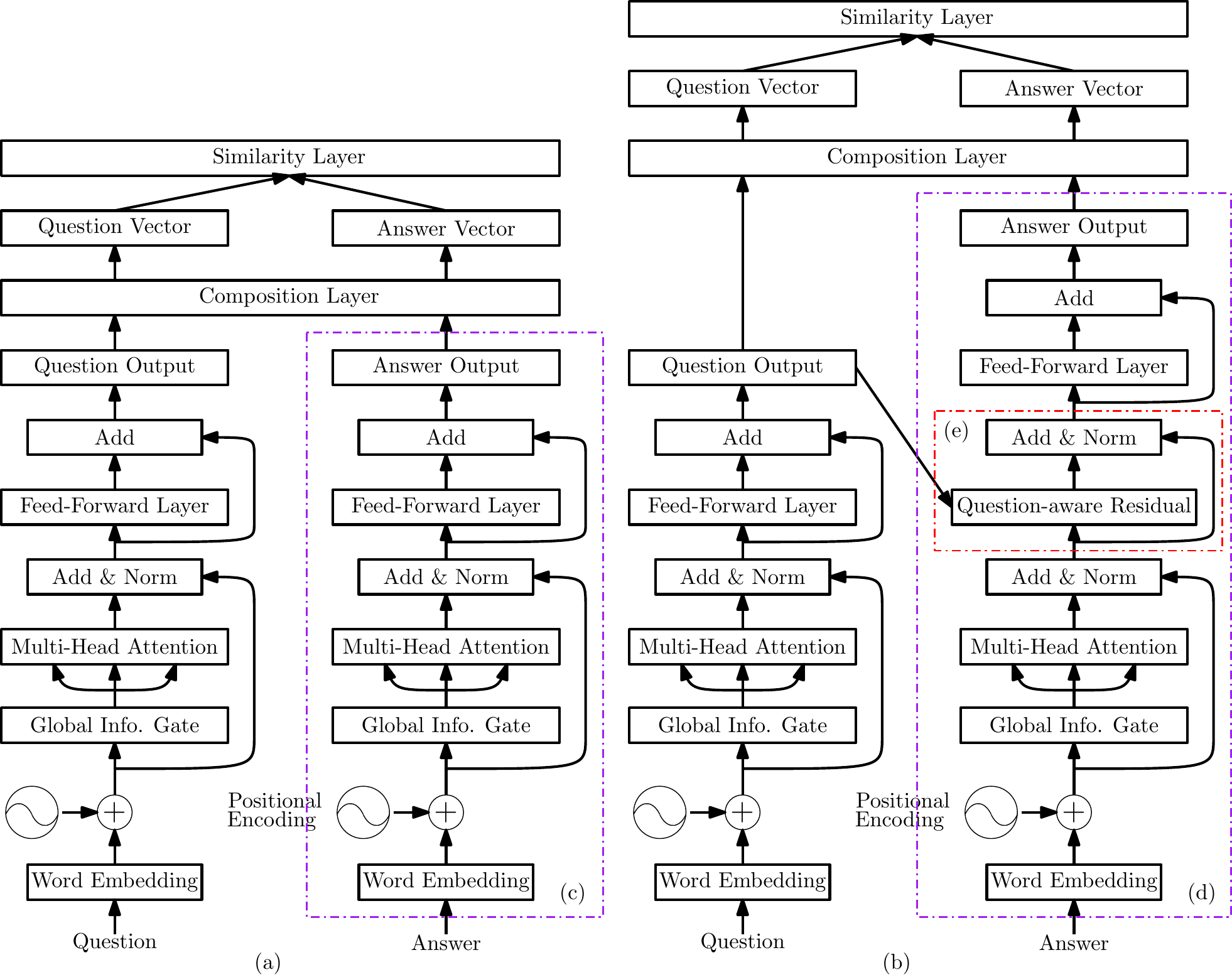}
    \caption{\label{GGSAN-i} (a). The architecture of GGSA based deep models for answer selection. (b). The architecture of iGGSA based deep models for answer selection. (c). The structure of a GGSA encoder. (d). The structure of a iGGSA encoder. (e). The structure of question-answer interaction mechanism.}
  \end{figure*}
  
  The architecture of iGGSA is shown in Figure~\ref{GGSAN-i}~(b), which consists of a GGSA encoder for question and a iGGSA encoder with question-answer interaction for answer~(Figure \ref{GGSAN-i}~(d)). The formulation of iGGSA is as follows:
  \begin{gather}
    \bm{c}^q = \frac{\sum_i \bm{h}_i^q}{L}, \label{pooling}\\
    \widetilde{\bm{R}} = FeedForwardLayer(\bm{Y}^a \odot \bm{c}^q), \nonumber \\
    \widetilde{\bm{Y}}^a = LayerNorm(\bm{Y}^a + \widetilde{\bm{R}}), \nonumber \\
    \bm{R} = FeedForwardLayer(\widetilde{\bm{Y}}^a), \label{GGSAN-i-4}\\
    \bm{H}^a = \widetilde{\bm{Y}}^a + \bm{R}. \label{GGSAN-i-5}
  \end{gather}
  where $\bm{h}^q_i \in \mathbb{R}^D$ is the $i$-th column of the output of GGSA for question $q$, $L$ is the length of the question. $\bm{c}^q$ denotes the result of a mean pooling on the words of $\bm{H}^q \in \mathbb{R}^{D \times L}$, which can be seen as the vector representation of the question. $\bm{Y}^a$ is the output of (\ref{middle}) in the answer part. $\widetilde{\bm{R}}$ is the question-answer interaction residual which is added to $\bm{Y}^a$. $\widetilde{\bm{Y}}^a$ is the intermediate representation of the answer, which captures the context of the question through residual. The left operations to get the final representation of the answer with interaction are computed by (\ref{GGSAN-i-4}) and (\ref{GGSAN-i-5}).
  
  That is to say, in GGSA, representations of both questions and answers are computed based on (\ref{GGSAN-6}). In iGGSA, representation of questions is computed based on (\ref{GGSAN-6}), while representation of answers is computed based on (\ref{GGSAN-i-5}).

  \subsection{On the Top of GGSA}
  GGSA or iGGSA can be seen as an encoder like LSTM. The output corresponding to an input sequence in both \mbox{GGSA} and iGGSA can be represented by a matrix $\bar{\bm{H}} \in \mathbb{R}^{D \times L}$. $\bar{\bm{H}} = \bm{H}$ for both question and answer in GGSA, while in iGGSA $\bar{\bm{H}} = \bm{H}$ for a question and $\bar{\bm{H}} = \bm{H}^a$ for an answer. If we want to use some similarity functions like cosine similarity to measure the similarity between questions and answers, $\bar{\bm{H}}$ must be composed into a vector. The output format in both \mbox{GGSA} and iGGSA is similar to that in bidirectional \mbox{LSTM}~(BLSTM), which is a sequence of vector representations. Hence, all the composition methods used in BLSTM, including pooling, attention and multihop sequential attention~\cite{tran2018multihop}, can also be used in GGSA and iGGSA. We call this part \emph{composition module}. After composition, we can get a vector representation for each question or answer.
  
  Then, these vector representations can be used to calculate the similarities of questions and answers for ranking. At training time, some \emph{loss functions} can be defined based on the vector representation of questions and answers. Here, we introduce two representative loss functions: pairwise loss function~\cite{tan2016lstm, tran2018multihop} and pointwise loss function~\cite{tay2018cross}.
  
  The pairwise loss function is defined as follows:
  \begin{equation}
    \mathcal{J} = \sum_{i=1}^N max(0, 0.1 - cos(\bm{v}^{q_i}, \bm{v}^{a_i^+}) + cos(\bm{v}^{q_i}, \bm{v}^{a_i^-})), \nonumber
  \end{equation}
  where $q_i$ is the $i$-th question in dataset, $a_i^+$ is the answer corresponding to $q_i$, and $a_i^-$ is a randomly selected answer irrelevant to $q_i$. $\bm{v}^{q_i}, \bm{v}^{a_i^+}, \bm{v}^{a_i^-}$ are their vector representations, respectively.
  This loss function has been the most common loss function in answer selection.
  
  The pointwise loss function is defined as follows:
  \begin{equation}
    \mathcal{J} = - \sum_{i=1}^N [y_i log s_i + (1 - y_i) log (1 - s_i)], \nonumber
  \end{equation}
  where $y_i \in \{0, 1\}$ is the label denoting whether current answer corresponds to the question or not and $s_i$ is the probability of current answer corresponding to the question generated by models. Pointwise loss is another common loss function in answer selection. To generate $s_i$, a Multilayer Perceptron~(MLP) and a Softmax layer should be used after composition module:
  \begin{equation}
    \bm{p} = Softmax( MLP([\bm{v}^{q_i}, \bm{v}^{a_i}]) ), \\ \nonumber
  \end{equation}
  \begin{equation}
    s_i = p_1, \\ \nonumber
  \end{equation}
  \begin{equation}
    1 - s_i = p_0, \\ \nonumber
  \end{equation}
  where $\bm{v}^{q_i}$ and $\bm{v}^{a_i}$ are the vector representations of a question $q_i$ and an answer $a_i$ respectively. $\bm{p} \in \mathbb{R}^2$ is the generated probability distribution. The second item $p_1 \in \mathbb{R}$ of $\bm{p}$ is the probability of answer $a_i$ corresponding to the question $q_i$, which is denoted as $s_i$. The first item $p_0$ is the probability of answer $a_i$ irrelevant to the question $q_i$, which is denoted as $1-s_i$.
  
  With GGSA and iGGSA as a key building block, we can adopt the above \emph{upper structures}, including composition module and loss functions, to design the whole question-answering model. The whole model is shown in Figure~\ref{GGSAN-i}. At test time, the cosine similarity of question-answer pairs or the corresponding probability $s_i$ is directly used for ranking.
  
  \subsection{Complexity Analysis}
  The computational complexity of each attention process is related to the size of receptive fields and the dimension of vector representations. In global self-attention, each word will attend to all words in the sequence. So, the computational complexity of multi-head attention is $O(n \times L \times L \times d) = O(L \times L \times D)$, where $L$ is the number of words in the sequence, $n$ is the number of heads, $d$ is the dimension of a single head and $n \times d = D$ is the dimension of word embedding. In \mbox{GGSA}, the sequence is divided into $m$ groups with group size $l$, where $m \times l = L$. Self-attention is performed within groups in different heads. So, the computational complexity of multi-head attention is $O(n \times m \times l \times l \times d) = O(L \times l \times D)$.
  
  Compared with global self-attention, our GGSA can reduce the computational complexity of multi-head attention by a factor of $m$. Hence, our GGSA will become much more practical than global self-attention to handle datasets with long sequences.

  \section{Experiment}
  GGSA and iGGSA can be seen as encoders like LSTM.  Hence, our experiments are designed to demonstrate that \mbox{GGSA} and iGGSA as encoders are superior to other encoders like \mbox{LSTM} and global self-attention on QA tasks. The upper structure on the top of GGSA and iGGSA is not the focus of this paper.
  \subsection{Datasets}
  We evaluate the effectiveness of our models on two popular answer selection datasets, FAQ dataset insuranceQA and \mbox{CQA} dataset yahooQA. The statistics of the datasets are listed in Table~\ref{datasets}.
  \begin{table}[h]
   \vspace{-5pt}
    \caption{Statistics of the datasets.}
    \label{datasets}
    \centering
    \vspace{-10pt}
    \begin{tabular}{|l|c|c|}
      \hline
      & insuranceQA & yahooQA \\
      \hline
      Train \#Q & 12887 & 50112 \\
      Dev \#Q  & 1000 & 6289 \\
      Test1 \#Q & 1800 & 6283 \\
      Test2 \#Q & 1800 & --- \\
      \hline
      \#A per Q & 500 & 5 \\
      \hline
      Vocabulary & 22353 & 102705 \\
      \hline
    \end{tabular}
  \end{table}
  
  \emph{insuranceQA} is a FAQ dataset from insurance domain~\cite{feng2015applying}. We use the version 1 of this dataset, which has been widely used in existing works~\cite{tan2016lstm,deng2018knowledge,tran2018multihop}. This dataset has already been partitioned into four subsets, Train, Dev, Test1 and Test2. The average length of the questions and answers are 7 and 95, respectively. The number of candidate answers, including positive and negative answers, for each question is 500. There is more than one positive answer to some questions. As in existing work~\cite{feng2015applying,tran2018multihop}, we adopt P@1~(Precision@1) as evaluation metric for this dataset.
  
  \emph{yahooQA}~\footnote{\url{https://webscope.sandbox.yahoo.com/catalog.php?datatype=l&guccounter=1}} is a CQA dataset collected from Yahoo! Answers. For fair comparison, we adopt the dataset splits as those in~\cite{tay2017learning,tay2018cross} where questions and answers are filtered by their length. More specifically, the sentences with length beyond the range of 5 - 50 are filtered out. There are five candidate answers to each question, in which only one answer is positive. The other four negative answers are sampled from the top $1000$ hits using Lucene search for each question. As in existing work~\cite{tay2017learning,tay2018cross}, we adopt P@1~(Precision@1) and MRR~(Mean Reciprocal Rank) as evaluation metrics for this dataset.
  
  \subsection{Hyperparameters and Baselines}
  All models use the 300-dimensional pre-trained word embeddings as input. The dimensions of the hidden layer in RNN~(LSTM) based models are consistent with the settings in existing works where dimension is $141$ for insuranceQA~\cite{tan2016lstm} and $256$ for yahooQA~\cite{tay2018cross}. The dimension of self-attention based models is consistent with the dimension of word embedding. All models are optimized using RMSProp with momentum. Initial learning rate is tuned among \{$1*e^{-4}, 5*e^{-5}, 2*e^{-5}, 1*e^{-5}$\}. Dropout is the main regularization way, which is performed on the word embeddings with keep probability being $0.7$ following~\cite{vaswani2017attention}. When it is used in self-attention models, dropout is performed on the word embeddings before positional encoding is added. The training epochs are chosen to achieve the best results on the validation set. All reported results are the average of $5$ runnings.
  
  There are also some hyperparameters of GGSA. The number of heads is set to $6$. Group size is set to $10$. We use offset $0$ for $3$ heads and offset $5$ for the other $3$ heads. The feed-forward layers are the same as those in~\cite{vaswani2017attention}. Some hyperparameters (e.g. offset) have not been finely tuned, so the results of GGSA may be further improved. The GGSA/iGGSA model relies more on the group size and the number of heads according to the group self-attention mechanism. We also conduct a sensitivity analysis of the number of heads and the group size in Section \ref{sensitivity}.
  
  The state-of-the-art baselines on the two datasets are different. Hence, we adopt different baselines for comparison on different datasets.
  
  CNN and CNN with GESD~\cite{feng2015applying}, QA-LSTM~\cite{tan2016lstm}, AP-LSTM~\cite{tan2016improved}, IARNN-GATE~\cite{wang2016inner}, Multihop-MLP-LSTM and Multihop-Sequential-LSTM~\cite{tran2018multihop} are adopted as baselines on \emph{insuranceQA}. Among them, Multihop-Sequential-LSTM is the state-of-the-art model without external knowledge. We also compare GGSA with global self-attention which is the core of Transformer~\cite{vaswani2017attention} and compare group self-attention with local self-attention. We use the pairwise loss function for insuranceQA to keep in line with existing works~\cite{feng2015applying,tan2016lstm,tan2016improved,wang2016inner,tran2018multihop}. Batch size is fixed to $128$ and max length of the sequence is $200$. Word embeddings are pre-trained on dataset and also updated during training.
  
  NTN-LSTM, HD-LSTM~\cite{tay2018hyperbolic}, AP-CNN and AP-BiLSTM~\cite{santos2016attentive}, and CTRN~\cite{tay2018cross} are adopted as baselines on \emph{yahooQA}. Among them, CTRN is the state-of-the-art model. Following the setting in~\cite{tay2018cross}, we use pointwise loss function for yahooQA. Before calculating loss, a two-layer MLP and a Softmax layer are applied to calculate the matching score. Batch size is $256$ and the sequence length is $50$. Word embeddings are initialized with pre-trained GloVe embeddings~\cite{pennington2014glove} and fixed during training, which are learned by a projection layer instead.
  
  \subsection{Experimental Results}
  \subsubsection{Effectiveness of GGSA Block}
  Firstly, we aim to prove that our proposed model GGSA as an encoder block can outperform LSTM and global self-attention. The comparison is performed in the cases of using max-pooling and attention as composition module. Table~\ref{exp1} shows the result. QA-LSTM~\cite{tan2016lstm} is the model using BLSTM as encoder for questions and answers, with a max-pooling operation for composition. Because \mbox{QA-LSTM} and global self-attention have no interaction between questions and answers before upper structure, we don't compare them with iGGSA. We can find that even GGSA can outperform these two baselines, which proves the effectiveness of GGSA. 
  \begin{table}[h]
    \caption{Comparison with LSTM and global self-attention on insuranceQA when using max-pooling or attention as composition module.}
    \vspace{-10pt}
    \label{exp1}
    \centering
    \begin{tabular}{|l|c|c|}
      \hline
      Model (insuranceQA) & Test1 & Test2 \\
      \hline
      QA-LSTM & 66.08 & 62.63 \\
      global self-attention with max-pooling & 67.61 & 64.03 \\
      GGSA with max-pooling  & \textbf{68.10} & \textbf{64.84} \\
      \hline
      QA-LSTM with attention & 71.10 & 66.96 \\
      global self-attention with attention & 70.38 & 67.30 \\
      GGSA with attention & \textbf{71.29} & \textbf{68.25} \\
      \hline
    \end{tabular}
  \end{table}
  
  We have introduced that both group self-attention and local self-attention can be adopted in our GGSA model. Here, we compare these two self-attention methods in details. The results are shown in Table~\ref{exp2-2}. Local self-attention represents the model that a word in the sequence only attends to the words in the sliding window.  The result of local self-attention is much lower than GGSA due to the lack of global information, which also proves the effectiveness of GGSA. We also incorporate the global information gate into local self-attention, which named local self-attention~(+gate). The result of local self-attention~(+gate) is comparable to GGSA, but with a much slower running speed. For efficiency consideration, we choose group self-attention as the key building block to design GGSA/iGGSA.
  \begin{table}[h]
    \caption{Comparison of group self-attention and local self-attention. All models use max-pooling for composition.}
    \vspace{-10pt}
    \label{exp2-2}
    \centering
    \begin{tabular}{|l|c|c|c|}
      \hline
      Model (insuranceQA) & Test1 & Test2 & Seconds / Batch\\
      \hline
      local self-attention & 67.23 & 64.59 & 1.28 \\
      local self-attention (+gate) & 67.90 & 65.95 & 1.30 \\
      GGSA  & 68.10 & 64.84 & 0.32 \\
      \hline
    \end{tabular}
  \end{table}
  
  \subsubsection{Effectiveness of Offset Strategy, Global Information Gate and Interaction Mechanism}
  We also prove the effectiveness of offset strategy, global information gate and interaction mechanism. The results on insuranceQA are shown in Table~\ref{exp2}. Group self-attention is the model in which the global multi-head attention is simply replaced by group multi-head attention without offset and global information gate. We can see that the accuracy of group self-attention is lower than that of global self-attention. This is reasonable because group self-attention itself does not have the ability to model global information. By adding offset strategy to group self-attention, the accuracy is improved. This improvement is due to the expansion of its receptive field, which proves the effectiveness of offset strategy. The accuracy can be further improved by adding global information gate, which constitutes our proposed model GGSA finally. GGSA outperforms global self-attention, which indicates that it is effective to distinguish local information from global information. This result also indicates that the global information gate can incorporate the global information into the representation of words, which proves that the motivation of global information gate is right. Furthermore, iGGSA gains an improvement of 3\% compared to GGSA, which proves the effectiveness of interaction mechanism proposed in this paper.
  \begin{table}[t]
    \caption{Effectiveness of offset, global information gate and interaction mechanism. All models use max-pooling for composition.}
    \vspace{-10pt}
    \label{exp2}
    \centering
    \begin{tabular}{|l|c|c|}
      \hline
      Model (insuranceQA) & Test1 & Test2 \\
      \hline
      global self-attention & 67.61 & 64.03 \\
      group self-attention & 66.52 & 63.15 \\
      group self-attention (+offset)& 67.09 & 63.25 \\
      GGSA (+offset,+gate)  & 68.10 & 64.84 \\
      iGGSA & \textbf{71.33} & \textbf{68.40} \\
      \hline
    \end{tabular}
  \end{table}

  \subsubsection{Results on insuranceQA}
  We compare GGSA and \mbox{iGGSA} with baselines on insuranceQA dataset. The results are shown in Table~\ref{exp3}. Note that the result of our implementation for Multihop-Sequential-LSTM~(our impl.) is better than the result from the original paper. This is due to some details, such as pretrained word embedding and hyperparameter tuning, are different. In order to ensure the fairness of comparison, we apply the same setting~\footnote{We find dropout on input can get more than one percent improvement for LSTM-based models on this dataset.} of GGSA to Multihop-Sequential-LSTM, which denoted as Multihop-Sequential-LSTM+~(our impl.). Besides max-pooling and attention, we also adopt multihop sequential attention as composition module for iGGSA, which is named iGGSA with multihop-seq-att. Multihop sequential attention is more effective than max-pooling and attention, which has already been proved in \cite{tran2018multihop}. We can see that it achieves the state-of-art result on this dataset, outperforming the counterpart with LSTM as an encoder. The results verify the effectiveness of our GGSA and iGGSA.
  
  \begin{table}[h]
    \caption{Results on insuranceQA. The results of models marked with $\star$ are reported from~\cite{tran2018multihop}.}
    \vspace{-10pt}
    \label{exp3}
    \centering
    \begin{tabular}{|l|c|c|}
      \hline
      Model & Test1 & Test2 \\
      \hline
      CNN $\star$ & 62.80 & 59.20 \\
      CNN with GESD $\star$ & 65.30 & 61.00 \\
      QA-LSTM (our impl.) & 66.08 & 62.63 \\
      AP-LSTM $\star$ & 69.00 & 64.80 \\
      IARNN-GATE $\star$ & 70.10 & 62.80 \\
      Multihop-MLP-LSTM $\star$ & 69.50 & 65.50 \\
      Multihop-Sequential-LSTM $\star$ & 70.50 & 66.90 \\
      Multihop-Sequential-LSTM~(our impl.) & 71.09 & 67.45 \\
      Multihop-Sequential-LSTM+~(our impl.) & 73.93 &  70.10 \\
      \hline
      GGSA with max-pooling  & 68.10 & 64.84 \\
      GGSA with attention & 71.29 & 68.25 \\
      iGGSA with max-pooling & 71.33 & 68.40 \\
      iGGSA with attention & 71.54 & 67.63 \\
      iGGSA with multihop-seq-att & \textbf{74.47} & \textbf{71.40} \\
      \hline
    \end{tabular}
  \end{table}

  \subsubsection{Results on yahooQA}
  We also evaluate our models and baselines on yahooQA dataset. Table~\ref{exp4} shows the results. We can see that our proposed models GGSA and iGGSA outperform the baselines, including the state-of-the-art baseline \mbox{CTRN},  by a large margin. Once again, the results on \mbox{yahooQA} verify the effectiveness of our GGSA and iGGSA.
  \begin{table}[t]
    \caption{Results on yahooQA. All models are optimized by pointwise loss. The results of models marked with $\star$ are reported from~\cite{tay2018cross}.}
    \vspace{-10pt}
    \label{exp4}
    \centering
    \begin{tabular}{|l|c|c|}
      \hline
      Model & P@1 & MRR \\
      \hline
      Random Guess & 20.00 & 45.86 \\
      NTN-LSTM $\star$ & 54.50 & 73.10 \\
      HD-LSTM $\star$ & 55.70 & 73.50 \\
      AP-CNN $\star$ & 56.00 & 72.60 \\
      AP-BiLSTM $\star$ & 56.80 & 73.10 \\
      CTRN $\star$ & 60.10 & 75.50 \\
      CTRN~(our impl.) & 60.71 & 77.26 \\
      \hline
      GGSA with max-pooling & 61.99 & 78.15 \\
      GGSA with attention & 65.16 & 80.26 \\
      iGGSA with max-pooling & 70.28 & 83.66 \\
      iGGSA with attention & \textbf{74.26} & \textbf{86.28}\\
      \hline
    \end{tabular}
  \end{table}

      \begin{figure*}[ht]
    \begin{minipage}{0.48\textwidth}
      \centering
      \includegraphics[width=0.9\columnwidth]{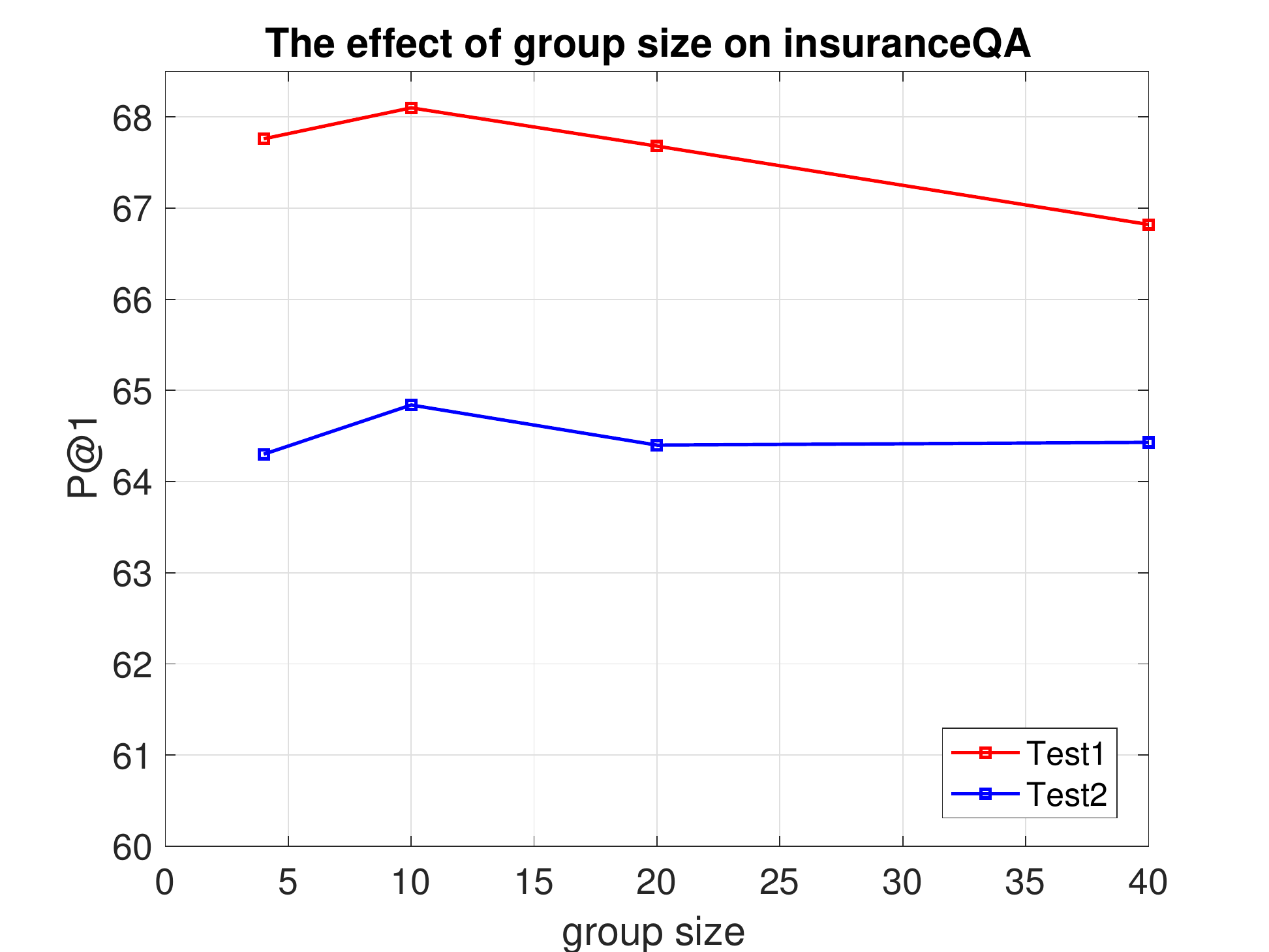}
    \end{minipage}
    \hfill
    \begin{minipage}{0.48\textwidth}
      \centering
      \includegraphics[width=0.9\columnwidth]{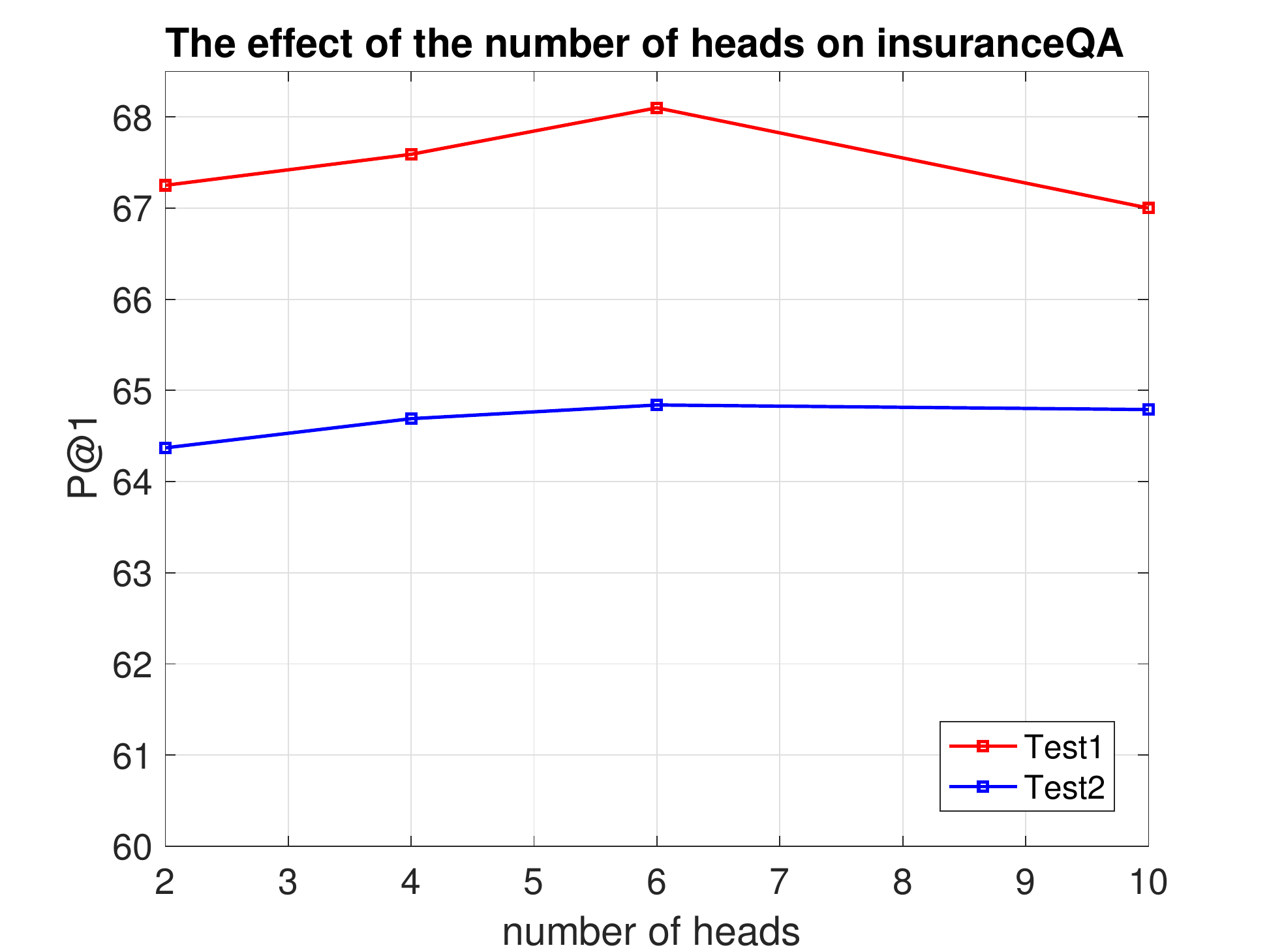}
    \end{minipage}
    \caption{\label{exp5} Sensitivity analysis of the group size and the number of heads. Each comparison is performed on both testsets (Test1 and Test2) on insuranceQA.}
  \end{figure*}
  \subsubsection{Sensitivity Analysis}
  \label{sensitivity}
  The GGSA/iGGSA model relies more on the group size and the number of heads according to the group self-attention mechanism. So we design a sensitivity study of these two important hyperparameters on insuranceQA in this section. Figure \ref{exp5} shows that GGSA is not sensitive to the group size in a proper range $[4, 20]$ on both testsets, Test1 and Test2. All three results in this range surpass the results of global self-attention with max-pooling, which are 67.24 and 63.98 on Test1 and Test2 respectively. The group size is not the bigger the better, since the bigger group size will conflict with the global information gate. Figure \ref{exp5} also shows the effect of the number of heads. More heads make the model attend to more different contents of the sequence theoretically, which does not lead to a better performance as the number of heads exceeds $6$. Since the dimension of hidden state is fixed, more heads lead to a smaller dimension of the subspaces, which limits the model capacity.

  The performance of GGSA is also sensitive to loss functions. We have introduced two loss functions before, which named pairwise loss and pointwise loss respectively. Here, we design a comparison of these two loss functions on yahooQA. The outputs of models play different roles by using different loss functions. For pairwise loss, the models generate the matching score of question-answer pairs for ranking. While for pointwise loss, the models generate the probability of current answer corresponding to the question. The comparison is shown in Table~\ref{exp4-2}. We can find that pairwise loss is better than pointwise loss for P@1 metric, while pointwise loss is superior than pairwise loss for MRR metric. Our experiments are mainly designed for comparing GGSA with other encoders like LSTM, rather than comparing upper structures. So the loss function adopted in each dataset is selected for fair comparison. With this comparison of these two loss functions, the state-of-the-art P@1 result on yahooQA can be further improved by using pairwise loss.
  
  \begin{table}[h]
    \caption{The comparison of pairwise loss and pointwise loss on yahooQA.}
    \vspace{-10pt}
    \label{exp4-2}
    \centering
    \begin{tabular}{|l|c|c|}
      \hline
      Model & P@1 & MRR \\
      \hline
      GGSA with max-pooling (pointwise loss) & 61.99 & 78.15 \\
      GGSA with max-pooling (pairwise loss) & 64.85 & 76.40 \\
      \hline
      iGGSA with attention (pointwise loss) & 74.26 & 86.28 \\
      iGGSA with attention (pairwise loss) & 78.45 & 85.07\\
      \hline
    \end{tabular}
  \end{table}

  \section{Related Work}
  First of all, our work is related to answer selection~(answer ranking). Existing models for answer selection can be classified into two categories, shallow~(non-deep) models and deep models. Shallow models usually use manually designed rules~\cite{riloff2012a}, dependency parse trees~\cite{wang2007jeopardy,wang2010probabilistic}, term-frequency inverse-document-frequency (TF-IDF) or bag-of-words~(BOW)~\cite{yih2013question} as sentence features. Then, distance measure functions are used to measure the similarity of questions and answers. All of these models are based on surface features and seldom consider semantics. The earliest deep models use CNN~\cite{feng2015applying} or LSTM~\cite{tan2016lstm} as the encoder block for questions and answers. Pooling is usually adopted as composition module to get the question and answer representations. For getting a better representation of the answer, attention mechanism borrowed from neural machine translation~(NMT)~\cite{bahdanau2015neural} is usually used to generate answer representation~\cite{tan2016lstm}. Stacked LSTMs are adopted in~\cite{wang2015a} to learn a joint feature vector of the question-answer pair. The following works are more concerned about the interaction between question-answer pairs. \cite{wan2016deep} calculates the word by word similarity and then a MLP is applied to similarity matrix to generate the final question-answer similarity. For uncovering the complex relations of question and answer pairs, \cite{tran2018multihop} proposes the strategy of multihop attention. \cite{wang2016inner} and \cite{chen2018rnn} introduce the representation of question to the answer encoder through gates, which is also called inner attention. \cite{tay2018cross} designs a CTRN cell to capture the interaction between question-answer pairs.
  
  Our work is also concerned with self-attention. Self-attention (global self-attention) is first proposed in~\cite{vaswani2017attention} for NMT and has been successfully applied in many applications. \cite{wang2018non} replaces CNN with self-attention to solve video classification problems. \cite{tan2018deep} applies self-attention in semantic role labeling task. \cite{zhou2018atrank} and \cite{sperber2018self} apply self-attention into recommendation model and acoustic model respectively. Self-attention has also been used in reading comprehension models~\cite{wang2017gated,wei2018fast}. To the best of our knowledge, self-attention~(with multi-head attention and residual structure) has not been adopted in answer selection.

  \section{Conclusion}
  In this paper, we propose a new self-attention based deep model, called GGSA, for answer selection. This is the first work to propose self-attention based deep models for answer selection. GGSA can distinguish global and local information by group self-attention. Offset strategy is used to enlarge receptive field, and global information gate is used to capture global information. We also propose a novel interaction mechanism with a residual structure to enhance GGSA. GGSA can outperform existing answer selection methods to achieve state-of-the-art performance on two QA datasets. Moreover, GGSA can also achieve higher accuracy than global self-attention for the answer selection task, with a lower computation cost.
  
  As a general encoder, GGSA can also be applied to other NLP tasks like reading comprehension, which will be pursued in our future work.

  \bibliographystyle{ACM-Reference-Format}
  \bibliography{ref}

\end{document}